\documentclass[11pt,a4paper]{article}
\usepackage[hyperref]{emnlp2020}
\usepackage{times}
\usepackage{latexsym}

\usepackage{microtype}

\aclfinalcopy 
\def\aclpaperid{96} 

\usepackage[linesnumbered, ruled]{algorithm2e}
\usepackage[T1]{fontenc}
\usepackage{dsfont}
\usepackage{colortbl}
\usepackage{xcolor}
\usepackage{enumitem}

\usepackage{mdframed}
\usepackage{booktabs}
\usepackage{graphicx}

\usepackage[normalem]{ulem}

\usepackage{xcolor}
\usepackage{soul}
\colorlet{soulred}{red!30}
\sethlcolor{soulred}%
\usepackage{amsmath,amsthm,amsfonts,amssymb,bm}

\usepackage{framed}
\usepackage{varioref}
\usepackage{float}
\usepackage{multirow}
\usepackage{dashrule}
\usepackage{url}
\usepackage{pifont}
\newcommand{\eat}[1]{\ignorespaces}
\newcommand{\xxcomment}[4]{\eat{#1}}
\newcommand{\xd}[1]{\xxcomment{blue}{X}{D}{#1}}

\title{Event Extraction by Answering (Almost) Natural Questions}

\author{Xinya Du \ {\normalfont and} \ Claire Cardie\\
  Department of Computer Science\\
  Cornell University \\
  Ithaca, NY, USA \\
  {\tt \{xdu, cardie\}@cs.cornell.edu} \\
  }

\date{}

\begin{document}
\maketitle
\begin{abstract}
The problem of event extraction requires detecting the event trigger and extracting its corresponding arguments.
Existing work in event argument extraction typically relies heavily on entity recognition as a preprocessing/concurrent step, causing the well-known problem of error propagation.
To avoid this issue, we introduce a new paradigm for event extraction by formulating it as a question answering (QA) task that extracts the event arguments in an end-to-end manner.
Empirical results demonstrate that our framework outperforms prior methods substantially; in addition, it is capable of extracting event arguments for roles not seen at training time (i.e., in a zero-shot learning setting).\footnote{Our code and question templates for the work are open sourced at \url{https://github.com/xinyadu/eeqa} for reproduction purpose.}
\end{abstract}

\section{Introduction}
\begin{figure}[t]
\centering
\includegraphics[scale=0.6]{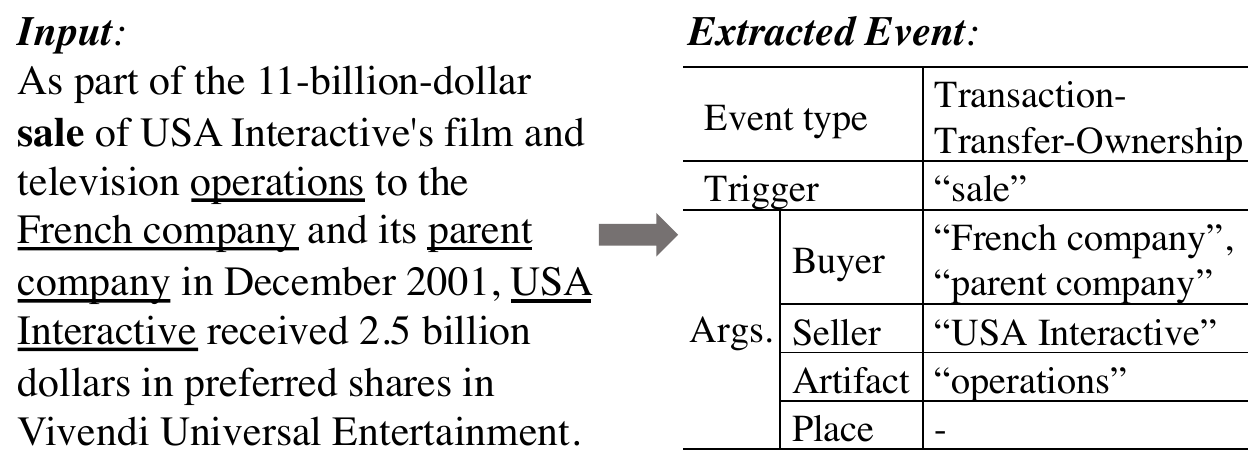}
\caption{Event extraction example from the ACE 2005 corpus~\cite{doddington2004automatic}.}
\label{fig:task}
\end{figure}

Event extraction is a long-studied and challenging task in Information Extraction
(IE)~\cite{sundheim1992overview, grishman-sundheim-1996-message, riloff1996automatically}. Its goal is to extract structured information --- ``what is happening'' and 
the persons/objects that are involved --- from unstructured text.
The task is illustrated via an example in Figure~\ref{fig:task},
which depicts an ownership transfer event (the \textit{event type}), triggered by the word ``sale" (the event \textit{trigger}) and accompanied
 by its extracted \textit{arguments} --- text spans denoting entities that
 fill a set of (semantic) \textit{roles} associated with the event type
 (e.g., \textsc{buyer}, \textsc{seller} and \textsc{artifact} for ownership transfer events).

Recent successful approaches to event extraction have benefited from dense features extracted by neural models~\cite{chen-etal-2015-event,nguyen-etal-2016-joint,liu-etal-2018-jointly} as well as contextualized lexical representations from pretrained language models~\cite{zhangtongtao2019joint,wadden-etal-2019-entity}.
The approaches, however, exhibit two key weaknesses.
First, they rely heavily on entity information for argument extraction. In particular, event argument extraction generally consists of two steps -- first identifying entities and their general semantic class
with trained models~\cite{wadden-etal-2019-entity} or a parser~\cite{sha2018jointly}, then assigning argument roles (or no role) to each entity.
Although joint models \cite{yang-mitchell-2016-joint,nguyen2019one,zhangjunchi2019extracting, lin-etal-2020-joint} have been proposed to mitigate this issue,
error propagation~\cite{li-etal-2013-joint} still occurs during event argument extraction.
\begin{figure*}[!t]
\centering
\resizebox{\textwidth}{!}{
\includegraphics[scale=0.65]{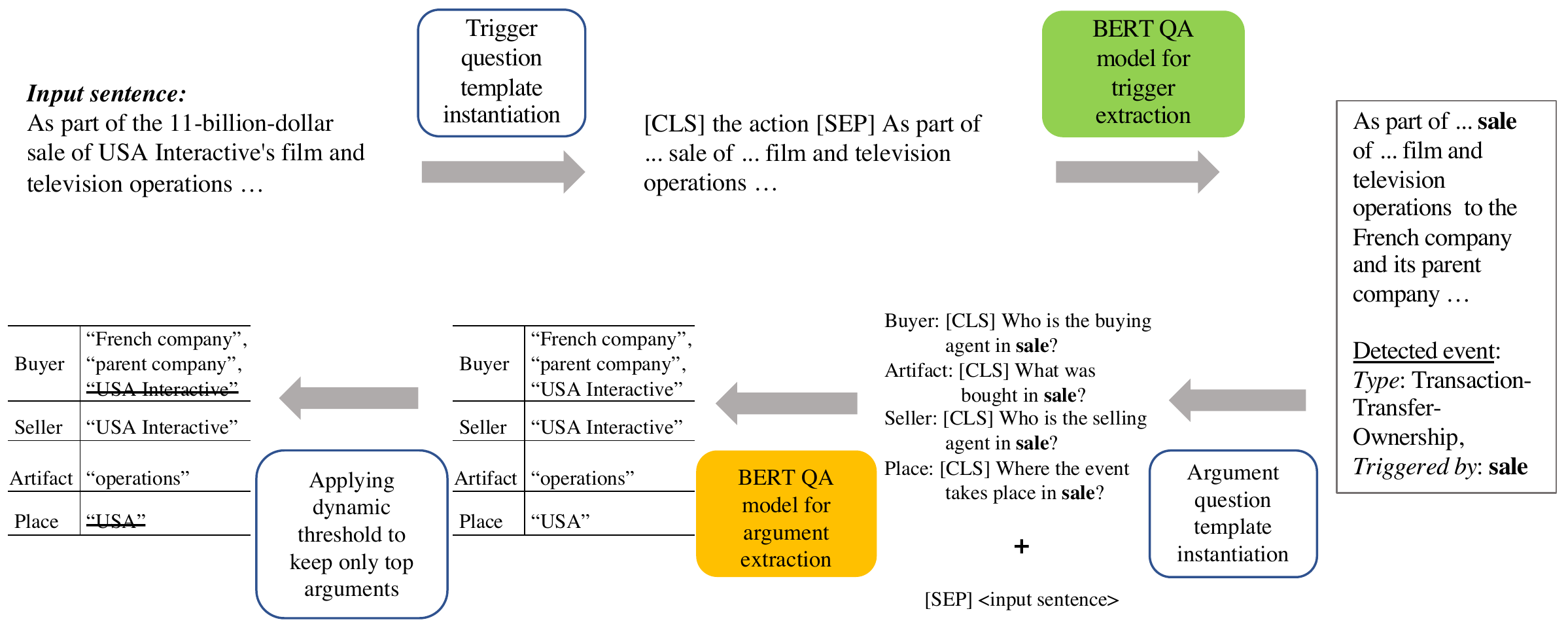}
}
\caption{Our framework for event extraction.}
\label{fig:framework}
\end{figure*}

A second weakness of neural approaches to event extraction is their inability to exploit the similarities of related argument roles across event types. For example, the ACE 2005~\cite{doddington2004automatic} \textsc{Conflict.Attack} events and \textsc{Justice.Execute} events have \textsc{target} and \textsc{person} argument roles, respectively. Both roles, however, refer to a \textit{human being (who) is affected} by an action. Ignoring the similarity can hurt performance, especially for argument roles with few/no examples at training time (e.g., similar to the zero-shot setting in~\newcite{levy-etal-2017-zero}).

In this paper, we propose a new paradigm for the event extraction task  -- formulating it as a question answering (QA)/machine reading comprehension (MRC) task (\textbf{Contribution 1}). 
The general framework is illustrated in Figure~\ref{fig:framework}. Using BERT~\cite{devlin-etal-2019-bert} as the
base model for obtaining contextualized representations from the input sequences, we develop two BERT-based QA models -- one for event trigger detection and the other for argument extraction. 
For each, we design one or more Question Templates that map the input sentence into the standard BERT input format. 
Thus, trigger detection becomes a request to identify ``the action'' or the ``verb'' in the input sentence and determine its event type; 
and argument extraction becomes a sequence of requests to identify the event's arguments, each of which is a text span in the input sentence.
Details will be explained in Section~\ref{sec:method}. To the best of our knowledge, this is the first attempt to cast event extraction as a QA task.

Treating event extraction as QA overcomes the weaknesses in existing methods identified above (\textbf{Contribution 2}):
(1) Our approach requires \textit{no entity annotation} (gold or predicted entity information) and no entity recognition pre-step; event argument extraction is performed as an end-to-end task;
(2) The question answering paradigm naturally permits the transfer of argument extraction knowledge across semantically related argument roles. 
We propose rule-based question generation strategies (including incorporating descriptions in annotation guidelines) for templates creation,
and conduct extensive experiments to evaluate our framework on the Automatic Content Extraction (ACE) event extraction task and  show empirically that the performance on both trigger and argument extraction outperform prior methods (Section~\ref{sec:results}).
Finally, we show that our framework extends to the zero-shot setting -- it is able to extract event arguments for unseen roles (\textbf{Contribution 3}).

\section{Methodology}
\label{sec:method}
In this section, we first provide an overview for the framework (Figure~\ref{fig:framework}), then go deeper into details of its components: question generation strategies for template creation, as well as training and inference of QA models.

\subsection{Framework Overview}
Our QA framework for event extraction relies on two sets of Question Templates that map an input sentence to a suitable input \textit{sequence} for two instances of a standard pre-trained bidirectional transformer (BERT~\cite{devlin-etal-2019-bert}). 
The first of these, BERT\_QA\_Trigger (green box in Figure~\ref{fig:framework}), extracts from the input sentence the event trigger which is a single token, and its type (one of a fixed set of pre-defined event types). 
The second QA model, BERT\_QA\_Arg (orange box in Figure~\ref{fig:framework}), is applied to the input sequence, the extracted event trigger and its event type to iteratively identify candidate event arguments (spans of text) in the input sentence.
Finally, a dynamic threshold is applied to the extracted candidate arguments, and only the arguments with probability above the threshold are retained.

\begin{table*}[t]
\centering
\scalebox{0.95}{
\begin{tabular}{l|lll}
\toprule
\multirow{2}{*}{Argument}      & Template 1  & Template 2 & Template 3  \\
  &  (Role name)  & (Type + Role question) & (Annotation guideline question) \\ \midrule          
Artifact           & artifact   & What is the artifact?    & What is being transported?                  \\
Agent              & agent        & Who is the agent?        & Who is responsible for the transport event? \\
Vehicle            & vehicle      & What is the vehicle?     & What is the vehicle used?                   \\
Origin             & origin        & What is the origination? & Where the transporting originated?          \\
Destination        & destination   & What is the destination? & Where the transporting is directed?  \\   
\bottomrule
\end{tabular}
}
\caption{Arguments (of event type \textsc{Movement.Transport}) and corresponding questions from three templates. ``in <trigger>'' is not added to the questions in this example.}
\label{tab:template_example}
\end{table*}

The input sequences for the two QA models share a standard BERT-style format:
\begin{equation}
\nonumber
\textbf{\text{[CLS] <question> [SEP] <sentence> [SEP]}}
\end{equation}
where [CLS] is BERT's special classification token, [SEP] is the special token to denote separation, and <sentence> is the tokenized input sentence.
We provide details on how to obtain the {\bf <question>} in Section~\ref{sec:strategies}. Details on the QA models and the inference process will be explained in Section~\ref{sec:models}.

\subsection{Question Generation Strategies}
\label{sec:strategies}

For our QA-based framework for event extraction to be easily moved from one domain to the other, we concentrated on developing question generation strategies that not only worked well for the task, but can be quickly and easily implemented.
For event trigger detection, we experiment with a set of four fixed templates -- ``what is the trigger'', ``trigger'', ``action'', ``verb''. Basically, we use the fixed literal phrase as the question. For example, if we choose the ``action'' template, the input sequence for the example sentence in Figures~\ref{fig:task} and~\ref{fig:framework} is instantiated as:
\begin{quote}
[CLS] action [SEP] As part of the 11-billion-dollar sale ... [SEP]
\end{quote}

As for event argument extraction, we design three templates with argument role name, basic argument based question and annotation guideline based question, respectively:
\begin{itemize} 
\item \textbf{Template 1 (Role Name)} For this template, <question> is simply instantiated with the argument role name (e.g., artifact, agent, place).

\item \textbf{Template 2 (Type + Role)} Instead of directly using the argument role name (<role name>) as the question, we first determine the argument role's general semantic type --- one of person, place, other; and construct the associated ``WH" word question -- \textit{who} for person, \textit{where} for place and \textit{what} for all other cases, of the following form:
\begin{quote}
\text{<WH\_word> is the <role name> ?}
\end{quote}
Examples are shown in Table 1 for the arguments of event type \textsc{Movement.Transport}. 
By adding the WH word, more semantic information is included as compared to Template 1.

\item \textbf{Template 3 (Incorporating Annotation Guidelines)} To incorporate even more semantic information and make the question more natural sounding, we utilize the descriptions of each argument role provided in the ACE annotation guidelines for events~\cite{ace-2005} for generating the questions.

\item \textbf{+ ``in <trigger>''} Finally, for each template type, it is possible to encode the trigger information by adding ``in <trigger>'' at the end of the question (where <trigger> is instantiated with the real trigger token obtained from the trigger detection phase). For example, the Template 2 question incorporating trigger information would be:
\begin{equation}
\nonumber
\text{<WH\_word> is the <argument> in <trigger>?}
\end{equation}

\end{itemize}

To help better understand all the strategies above, Table~\ref{tab:template_example} presents an example for argument roles of event type \textsc{Movement.Transport}. We see that the annotation guideline based questions are more natural and encode more semantics about a given argument role, than the simple Type + Role question ``what is the artifact?''.

\subsection{Question Answering Models}
\label{sec:models}

We use BERT~\cite{devlin-etal-2019-bert} as the base model for getting contextualized representations for the input sequences for both BERT\_QA\_Trigger and BERT\_QA\_Arg. 
After the instantiation with question templates the sequences are of format [CLS] <question> [SEP] <sentence> [SEP].

Then we get the contextualized representations of each token for trigger detection and argument extraction with $\texttt{BERT}_{Tr}$ and $\texttt{BERT}_{Arg}$, respectively.
For the input sequence $(e_1, e_2, ..., e_N)$ prepared for trigger detection, we have:
\begin{equation}
\begin{gathered}
\nonumber
\mathbf{E} = [\mathbf{e}_1, \mathbf{e}_2, ..., \mathbf{e}_N] \\
\mathbf{e}_1, \mathbf{e}_2, ..., \mathbf{e}_N = \texttt{BERT}_{Tr} (e_1, e_2, ..., e_N)
\end{gathered}
\end{equation}

For the input sequence $(a_1, a_2, ..., a_M)$ prepared for argument span extraction, we have:
\begin{equation}
\nonumber
\begin{gathered}
\mathbf{A} = [\mathbf{a}_1, \mathbf{a}_2, ..., \mathbf{a}_M] \\
\mathbf{a}_1, \mathbf{a}_2, ..., \mathbf{a}_M = \texttt{BERT}_{Arg} (a_1, a_2, ..., a_M)
\end{gathered}
\end{equation}

The output layer of each QA model, however, differs: 
BERT\_QA\_Trigger predicts the event type for each token in sentence (or None if it is not an event trigger), 
while BERT\_QA\_Arg predicts the start and end offsets for the argument span with a different decoding strategy.

More specifically, for \textbf{trigger prediction}, we introduce a new parameter matrix $\mathbf{W}_{tr} \in \mathbb{R}^{H \times T}$, where $H$ is the hidden size of the transformer and $T$ is the number of event types plus one (for non-trigger tokens). 
\texttt{softmax} normalization is applied across the $T$ types to produce $P_{tr}$, the probability distribution across the event types:
\begin{equation}
\nonumber
P_{tr} = \texttt{softmax} (\mathbf{E} \mathbf{W}_{tr}) \in \mathbb{R}^{T} \times N
\end{equation}
At test time, for trigger detection, to obtain the type for each token $e_1, e_2, ..., e_N$, we simply apply argmax to $P_{tr}$.

For \textbf{argument span prediction}, we introduce two new parameter matrices $\mathbf{W}_s \in \mathbb{R}^{H \times 1}$ and $\mathbf{W}_e \in \mathbb{R}^{H \times 1}$; softmax normalization is then applied 
across the input tokens $a_1, a_2, ..., a_M$ 
to produce the probability of each token being selected as the start/end of the argument span:
\begin{equation}
\begin{gathered}
\nonumber
P_{s} (i) = \texttt{softmax} (\mathbf{a}_i \mathbf{W}_s) \\
P_{e} (i) = \texttt{softmax} (\mathbf{a}_i \mathbf{W}_e)
\end{gathered}
\end{equation}

To train the models (BERT\_QA\_Trigger and BERT\_QA\_Arg), we minimize the negative log-likelihood loss for both models, parameters are updated during the training process. In particular, the loss for the argument extraction model is the sum of two parts: the start token loss and end end token loss.
For the training examples with no argument span (no answer case), we minimize the start and end probability of the first token of the sequence ([CLS]).
\begin{equation}
\begin{gathered}
\nonumber
    \mathcal{L}_{arg} = \mathcal{L}_{arg\_start} + \mathcal{L}_{arg\_end}
\end{gathered}
\end{equation}

\paragraph{Inference with Dynamic Threshold for Argument Spans}
At test time, predicting the argument spans is more complex -- for each argument role, there can be \textit{several} or \textit{no} spans to be extracted.
After the output layer, we have the probability of each token $a_i \in (a_1, a_2, ..., a_M)$ being the start ($P_{s}(i)$) and end ($P_{e}(i)$) of the argument span.
\begin{algorithm}[h]
\small
\SetKwData{maxlen}{MaxSpanLength}\SetKwData{This}{this}\SetKwData{Up}{up}
\SetKwInOut{Input}{Input}\SetKwInOut{Output}{Output}
  \Input{$P_{s}(i)$, where $i \in \{1,...,M\}$, \\
         $P_{e}(i)$, where $i \in \{1,...,M\}$}
  \Output{valid candidate spans for the argument role}
  \BlankLine
  \For{$start\leftarrow 1$ \KwTo $M$}{
    \For{$end\leftarrow 1$ \KwTo $M$}{
        \lIf{$start$ {\bf or} $end$ not in the input sentence}{continue}
        \lIf{$end-start+1 >$ MaxSpanLength}{continue}
        \lIf{$P_{s}(start) < P_{s}([CLS])$ {\bf or} $P_{e}(end) < P_{e}([CLS])$}{continue}
        \tcp{add the valid candidate span to the set}
        $score \leftarrow P_{s}(start) + P_{e}(end)$\;
        $no\_ans\_score \leftarrow P_{s}(1) + P_{e}(1) - score$\;
        $candidates.$add$([start, end, no\_ans\_score])$
    }
  }
\caption{Harvesting Argument Spans Candidates}
 \label{algo:argument}
\end{algorithm}
\begin{algorithm}[!h]
\small
\SetKwData{maxlen}{MaxSpanLength}\SetKwData{This}{this}\SetKwData{Up}{up}
\SetKwInOut{Input}{Input}\SetKwInOut{Output}{Output}
  \Input{$dev\_candidates(i)$, $i \in \{1,...,dev\_n\}$,\\
         $test\_candidates(i)$, $i \in \{1,...,test\_n\}$.}
  \Output{A set of top arguments from test\_candidates}
  \BlankLine
  \tcp{get the best dynamic threshold}
  sort$(dev\_candidates, key=no\_ans\_score)$\;
  $best\_thresh \longleftarrow 0$\;
  $best\_res \longleftarrow 0$\;
  \For{$i\leftarrow 1$ \KwTo $dev\_n$}{
     $thresh \leftarrow dev\_candidates(i).no\_ans\_score$\;
     $result \leftarrow$ eval($dev\_candidates$ with $no\_ans\_score <= thresh)$\;
     \lIf{$result > best\_res$}
     {$best\_thresh \leftarrow thresh;$\\
      \ \ $best\_res \leftarrow result$}
  }
  \tcp{apply the best threshold}
  $final\_arguments \longleftarrow \{\}$\;
  \For{$i\leftarrow 1$ \KwTo $test\_n$}{
    \lIf{$test\_candidates(i).no\_ans\_score <= best\_thresh$}
    {$final\_arguments.$add$(test\_candidates(i))$}
  }
\caption{Automatic Filtering on Argument Candidates}
\label{algo:argument2}
\end{algorithm}

Firstly, we run an algorithm to harvest all valid argument spans candidates for each argument role (Algorithm~\ref{algo:argument}). Basically, we:
\begin{enumerate}
    \item Enumerate all the possible combinations of start offset ($start$) and end offset ($end$) of the argument spans (line 1--2);
    \item Eliminate the spans not satisfying the constraints: start and end token must be within the sentence; 
    the length of the span should be shorter than a maximum length constraint;
    Argument spans should have larger probability than the probability of ``no argument'' (which is stored at the [CLS] token) (line 3--5);
    \item Calculate the relative no answer score ($no\_ans\_score$) for the candidate span and add the candidate to list (line 6--8).
\end{enumerate}

Then we run another algorithm to filter out candidate arguments that should not be included (Algorithm~\ref{algo:argument2}).
More specifically, we obtain a probability threshold ($best\_thresh$) that helps achieve best evaluation results on the dev set (line 1--9) and keep only those arguments with $no\_ans\_score$ smaller than the threshold (line 10--13). With the dynamic threshold for determining the number of arguments to be extracted for each role\footnote{Each role has a separate threshold.}, we avoid adding a (hard) hyperparameter for this purpose. 

Another easier way to get final argument predictions is to directly include all the candidates with $no\_ans\_score<0$, which does not require tuning the dynamic threshold $best\_thresh$.

\begin{table*}[t]
\centering
\resizebox{0.9\textwidth}{!}{
\begin{tabular}{l|ccc|ccc}
\toprule
& \multicolumn{3}{c|}{Trigger Identification} & \multicolumn{3}{c}{Trigger ID + Classification} \\ \midrule
     & P            & R            & F1           & P              & R              & F1            \\ \midrule
dbRNN~\cite{sha2018jointly} & - & - & - & 74.10 & 69.80 & 71.90 \\
Joint3EE~\cite{nguyen2019one}   & 70.50        & 74.50        & 72.50        & 68.00          & 71.80          & 69.80         \\
GAIL-ELMo~\cite{zhangtongtao2019joint} & 76.80        & 71.20        & 73.90        & 74.80          & 69.40          & 72.00         \\
DYGIE++, BERT + LSTM~\cite{wadden-etal-2019-entity} & -            & -            & -            & -              & -              & 68.90         \\
DYGIE++, BERT FineTune~\cite{wadden-etal-2019-entity}   & -            & -            & -            & -              & -              & 69.70         \\ \midrule
Our BERT FineTune   & 69.77        & 76.18        & 72.84        & 67.15          & 73.20          & 70.04         \\
BERT\_QA\_Trigger (best trigger question strategy) & 74.29        & 77.42        & 75.82        & 71.12          & 73.70          & \textbf{72.39}  \\ \bottomrule
\end{tabular}
}
\caption{Trigger detection results.}
\label{tab:result_trigger}
\end{table*}
\section{Experiments}

\subsection{Dataset and Evaluation Metric}
We conduct experiments on the ACE 2005 corpus~\cite{doddington2004automatic}, it contains documents crawled between year 2003 and 2005 from a variety of areas such as newswire (nw), weblogs (wl), broadcast conversations (bc) and broadcast news (bn). The part that we use for evaluation is fully annotated with 5,272 event triggers and 9,612 arguments. We use the same data split and pre-processing step as in the prior works~\cite{zhangtongtao2019joint,wadden-etal-2019-entity}.

As for evaluation, we adopt the same criteria defined in~\newcite{li-etal-2013-joint}:
An event trigger is correctly identified (ID) if its offsets match those of a gold-standard trigger; and it is correctly classified if its event type (33 in total) also matches the type of the gold-standard trigger.
An event argument is correctly identified (ID) if its offsets and event type match those of any of the reference argument mentions in the document; and it is correctly classified if its semantic role (22 in total) is also correct.
Though our framework does not involve the trigger/argument identification step and tackles the identification + classification in an end-to-end way, we still report the trigger/argument identification's results to compare to prior work. It could be seen as a more lenient evaluation metric, as compared to the final trigger detection and argument extraction metric (ID + Classification), which requires both the offsets and the type to be correct.
All the aforementioned elements are evaluated using precision (denoted as P), recall (denoted as R) and F1 scores (denoted as F1).

\begin{table*}[t]
\centering
\resizebox{\textwidth}{!}{
\begin{tabular}{l|ccc|ccc}
\toprule
& \multicolumn{3}{c}{Argument Identification}                              & \multicolumn{3}{|c}{Argument ID + Classification}                   \\ \midrule
& P                          & R                    & F1                   & P                    & R                    & F1                   \\ \midrule
dbRNN~\cite{sha2018jointly} & -                          & -                    & 57.20                & -                    & -                    & 50.10                \\
Joint3EE~\cite{nguyen2019one} & -                          & -                    & -       & 52.10                & 52.10                & 52.10                \\
GAIL-ELMo~\cite{zhangtongtao2019joint} & 63.30                      & 48.70                & 55.10                & 61.60                & 45.70                & 52.40 \\
DYGIE++, BERT + LSTM~\cite{wadden-etal-2019-entity} & -                          & -                    & 54.10                & -                    & -                    & 51.40                \\
DYGIE++, BERT + LSTM ensemble~\cite{wadden-etal-2019-entity}   & -                          & -                    & 55.40                & -                    & -                    & 52.50                \\ \midrule
BERT\_QA\_Arg (annot. guideline question template) & 58.02                      & 50.69                & 54.11                & 56.87                & 49.83                & 53.12$^*$  \\
\ \ \ \ \   w/o dynamic threshold  & 53.39  & 54.69  & 54.03 & 50.81 & 52.78 & 51.77 \\
BERT\_QA\_Arg (ensemble argument question template 2\&3)  & 58.90 &52.08	&55.29	&56.77	& 50.24	& \textbf{53.31}\\ \bottomrule
\end{tabular}
}
\caption{Argument extraction results. $^*$ indicates statistical significance (p < 0.05).}
\label{tab:result_argument}
\end{table*}

\subsection{Results}
\label{sec:results}

\paragraph{Evaluation on ACE Event Extraction}

We compare our framework's performance to a number of prior competitive models:
{\bf dbRNN}~\cite{sha2018jointly} is an LSTM-based framework that leverages the dependency graph information to extract event triggers and argument roles.
{\bf Joint3EE}~\cite{nguyen2019one} is a multi-task model that performs entity recognition, trigger detection and argument role assignment by shared BiGRU hidden representations.
{\bf GAIL}~\cite{zhangtongtao2019joint} is an ELMo-based model that utilizes a generative adversarial network to help the model focus on harder-to-detect events.
{\bf DYGIE++}~\cite{wadden-etal-2019-entity} is a BERT-based framework that models text spans and captures within-sentence and cross-sentence context.
\xd{change}{\bf OneIE}~\cite{lin-etal-2020-joint} is a joint neural model for extraction with global features.\footnote{Slightly different from our and \newcite{wadden-etal-2019-entity}'s data pre-processing, OneIE skips lines before the <text> tag (e.g., headline, datetime).}

In Table~\ref{tab:result_trigger}, we present the comparison of models' performance on trigger detection. 
We also implement a BERT fine-tuning baseline and it reaches nearly same performance as its counterpart in DYGIE++.
We observe that our BERT\_QA\_Trigger model with the best trigger questioning strategy reaches comparable (better) performance with the baseline models.\footnote{Note that OneIE is concurrent to our work and reports better performance. On trigger detection, it reaches 74.7 F1 as compare to our 72.39. On argument extraction (affected by trigger detection), it reaches 56.8 as compared to our 53.31.}

Table~\ref{tab:result_argument} shows the comparison between our model and baseline systems on argument extraction. Notice that the performance of argument extraction is directly affected by trigger detection. Because argument extraction correctness requires the trigger to which the argument refers to be correctly identified and classified.
We observe,
(1) Our BERT\_QA\_Arg model with the best argument question generation strategy (annotation guideline based questions) outperforms prior work significantly, although it uses no entity recognition resources;
(2) Drop of F1 performance from argument identification (correct offset) to argument ID + classification (both correct offset and argument role) is only around 1\%, while the gap is around 3\% for prior models which rely on entity recognition and a multi-step process for argument extraction. This once again demonstrates the benefit of our new formulation for the task as question answering.

To better understand how the dynamic threshold is affecting our framework's performance. We conduct an ablation study on this (Table~\ref{tab:result_argument}) and find that the threshold increases the precision and the general F1 substantially.
The last row in the table shows the test time ensemble performance of the predictions from BERT\_QA\_Arg trained with template 2 question, and another BERT\_QA\_Arg trained with template 3 question (the two relatively better questioning strategies).
The ensemble system outperforms the non-ensemble system in both precision and recall, demonstrating the benefit from both templates.

\paragraph{Evaluation on Unseen Argument Roles}
To verify how our formulation provides advantages for extracting arguments with unseen argument roles (similar to the zero-shot relation extraction setting in~\newcite{levy-etal-2017-zero}), we conduct another experiment, where we keep 80\% of the argument roles (16 roles) seen at training time, and 20\% (6 roles) only seen at test time. Specifically, the unseen roles are ``Vehicle, Artifact, Target, Victim, Recipient, Buyer''.
Notice that during training, we use the subset of sentences from the training set, which are known to contain arguments of seen roles as positive examples. At test time, we evaluate the models on the subset of sentences from the test set, which contains arguments of unseen roles.\footnote{We omit the trigger detection phase in this evaluation.}

\begin{table*}[t]
\small
\resizebox{\textwidth}{!}{
\begin{tabular}{l|cccccc|cccccc}
\toprule
& \multicolumn{6}{c|}{Predicted Triggers} & \multicolumn{6}{c}{Gold Triggers}    \\
\cmidrule{2-7} \cmidrule{8-13}
 & \multicolumn{3}{c}{Argument Identification} & \multicolumn{3}{c|}{Argument ID + C} & \multicolumn{3}{c}{Argument Identification} & \multicolumn{3}{c}{Argument ID + C} \\ \midrule
Question & P             & R            & F1           & P              & R              & F1             & P             & R            & F1           & P              & R              & F1    \\ \midrule
Role name & 47.50         & 51.22        & 49.29        & 44.85          & 48.78          & 46.74          & 56.12         & 67.01        & 61.09        & 51.95          & 63.19          & 57.02          \\
\ \ \ \ + in {<}trigger{>}                                                             & 53.86         & 51.91        & 52.87        & 51.63          & 50.17          & 50.89          & 69.00         & 64.76        & 66.81        & 64.70          & 61.28          & 62.94          \\ \midrule
Type + Role question & 51.02         & 47.74        & 49.33        & 48.64          & 45.83          & 47.19          & 60.31         & 62.15        & 61.22        & 57.17          & 59.20          & 58.17          \\
\ \ \ \ + in {<}trigger{>}                                                             & 54.61         & 50.69        & 52.58        & 52.98          & 48.96          & 50.89          & 70.38         & 62.85        & 66.40        & 67.55          & 60.59          & 63.88          \\ \midrule
Annot. guideline question & 51.17         & 51.22        & 51.19        & 48.99          & 49.83          & 49.40          & 60.03         & 68.40        & 63.94        & 57.08          & 65.97          & 61.21          \\
\ \ \ \ + in {<}trigger{>}                                                             & 58.02         & 50.69        & 54.11        & 56.87          & 49.83          & \textbf{53.12}          & 71.17         & 65.45        & 68.19        & 67.88          & 63.02          & \textbf{65.36} \\ \bottomrule        
\end{tabular}
}
\caption{Influence of question generation strategies on argument extraction.}
\label{tab:question_influence_arg}
\end{table*}

\begin{table}[t]
\small
\resizebox{\columnwidth}{!}{
\begin{tabular}{l|ccc}
\toprule
& \multicolumn{3}{c}{Argument ID + Classification} \\ \midrule
& P               & R              & F1            \\ \midrule
Random NE       & 26.61           & 24.77          & 25.66         \\
\begin{tabular}[c]{@{}l@{}}GAIL\\\cite{zhangtongtao2019joint}\end{tabular} & -          & -           & -         \\ \midrule
Our model   &                 &                &    \\
\ \ \ \ w/ Role name & 73.83           & 53.21          & 61.85  \\
\ \ \ \ w/ Type + Role Q & 77.18           & 55.05          & 64.26  \\
\ \ \ \ w/ Annot. Guideline Q  & 78.52           & 59.63          & \textbf{67.79} \\ \bottomrule
\end{tabular}
}
\caption{Evaluation on sentences containing unseen argument roles.}
\label{tab:result_unseen}
\end{table}

 Table~\ref{tab:result_unseen} presents the results. {\bf Random NE} is our random baseline that selects a named entity in the sentence, it has a reasonable performance of near 25\%. Prior models such as {\bf GAIL} are not capable of handling the unseen roles. \xd{change}\textbf{ZSTE}~\cite{huang-etal-2018-zero} is a framework for zero-shot transfer learning of event extraction with AMR. It maps each parsed candidate span to a specific type in a target event ontology. Its argument extraction results are affected by AMR performance and their reported F1 is around 20-30\% in their evaluation setting.

Using our QA-based framework, as we leverage more semantic information and naturalness into the question (from question template 1 to 2, to 3), both the precision and recall increase substantially.

\section{Further Analysis}

\subsection{Influence of Question Templates}
To investigate how the question generation strategies affect the performance of event extraction, we perform experiments on trigger and argument extractions with different strategies, respectively.

In Table~\ref{tab:question_influence_trigger}, we try different fixed questions for trigger detection. By ``leaving empty'', we mean instantiating the {\bf question} with empty string.\footnote{In this case, the model degrades to a token classification model, which matches our BERT FineTune baseline's performance.} There's no substantial gap between different alternatives. By using ``verb'' as the question, our BERT\_QA\_Trigger model achieves best performance (measured by F1 score). \xd{change}The QA model also encodes the semantic \textit{interactions} between the fixed question (``verb'') and the sentence, this explains why BERT\_QA\_Trigger is better than BERT FineTune in trigger detection.

The comparison between different question generation strategies for argument extraction is even more interesting. In Table~\ref{tab:question_influence_arg}, we present the results in two settings: event argument extraction with predicted triggers (the same setting as in Table~\ref{tab:result_argument}), and with gold triggers. In summary, we find that:
\begin{itemize} 
    \item \textit{Adding ``in <trigger>'' after the question consistently improves the performance.} It serves as an indicator for what/where the trigger is in the input sentence. Without adding the ``in <trigger>'', for each template (1, 2 \& 3), the F1 of models' predictions drop around 3 percent when given predicted triggers, and more when given gold triggers.
    \item \textit{Our template 3 questioning strategy which is most natural achieves the best performance.} As we mentioned earlier, template 3 questions are based on descriptions for argument roles in the annotation guideline, thus encoding more semantic information about the role name. And this corresponds to the accuracy of models' predictions -- template 3 is more effective than templates 1\&2 in both with ``in <trigger>'' and without ``in <trigger>'' settings.
    What's more, we observe that template 2 (adding a WH\_word to form the questions) achieves better performance than the template 1 (directly using argument role name).
\end{itemize}

\begin{table}[t]
\small
\centering
\begin{tabular}{l|ccc}
\toprule
& \multicolumn{3}{c}{Trigger ID + Classification} \\ \midrule
    & P              & R              & F1 \\ \midrule
leaving empty                     & 67.15          & 73.20          & 70.04 \\ 
``what is the trigger'' & 70.15          & 69.98          & 70.06 \\ 
``what happened''  & 70.53          & 69.48          & 70.00 \\ 
``trigger''                           & 69.73          & 71.46          & 70.59 \\ 
``action''                            & 72.25          & 71.71          & 71.98 \\
``verb''                              & 71.12          & 73.70          & \textbf{72.39} \\ \bottomrule
\end{tabular}
\caption{Effect of questioning on trigger detection.}
\label{tab:question_influence_trigger}
\end{table}

\subsection{Error Analysis}
We further conduct error analysis and provide a number of representative examples.
Table~\ref{tab:error} summarizes error statistics for trigger detection and argument extraction.

For event triggers, the majority of the errors relate to missing or spurious predictions, and only 8.29\% involve misclassified event types (e.g., an \textsc{Elect} event is mistaken for a \textsc{Start-Position} event).
For event arguments, on the sentences that come with at least one event in gold data, our framework extracts more arguments only around 14\% of the cases. Most of the time (54.37\%), our framework extracts fewer arguments than it should; this corresponds to the results in Table~\ref{tab:result_argument}, where the precision of our models are higher. In around 30\% of the cases, our framework extracts the same number of arguments as in the gold data, almost half of which match exactly the gold arguments.

\begin{table}[t]
\small
\centering
\begin{tabular}{c|c|c}
\toprule
Missing & Spurious & Wrong Type \\ \midrule
 46.08\%       & 	45.62\%        & 8.29\% \\ 
\bottomrule
\end{tabular}
\BlankLine
\BlankLine
\BlankLine
\BlankLine
\begin{tabular}{c|c|c|c}
\toprule
\multicolumn{2}{c|}{same number} & \multirow{2}{*}{more} & \multirow{2}{*}{less} \\
exact match  & not exact match  &      &      \\ \midrule
14.48\% & 17.21\% & 13.93\% & 54.37\% \\ 
\bottomrule
\end{tabular}
\caption{Trigger errors (upper table) and argument errors (lower table).}
\label{tab:error}
\end{table}

After examining the example predictions, we find that reasons for errors can be mainly divided into the following categories:
\begin{itemize}[leftmargin=*,noitemsep]

    \item \textit{More complex sentence structures.}
    In the following example, the input sentence has multiple clauses, each with trigger and arguments (such as when triggers are partial or elided). Our model is capable of also extracting ``Tom'' as another \textsc{Entity} of the \textsc{Contact.Meet} event in the first example:
    \begin{quote}
    [She]\textsubscript{\textsc{Entity}} \textbf{visited} the store and [Tom]\textsubscript{\textsc{Entity}} did too.
    \end{quote}
 
    But in the second example, when there is a higher-order event expressed spanning events in nested clauses, our model did not extract the entire \textsc{Victim} correctly, which shows the difficulty of handling complex clause structures.
    \begin{quote}
    Canadian authorities arrested two Vancouver-area men on Friday and charged them in the \textbf{deaths} of [329 passengers and crew members of an Air-India Boeing 747 that blew up over the Irish Sea in 1985, en route from Canada to London]\textsubscript{\textsc{Victim}}.
    \end{quote}

    \item \textit{Lack of reasoning with document-level context.} In the sentence ``MCI must now seize additional assets owned by Ebbers, to secure the \textbf{loan}.'' There is a \textsc{Transfer-Money} event triggered by loan, with MCI as the \textsc{Giver} and Ebbers, the \textsc{Recipient}. In the previous paragraph, it's mentioned that ``Ebbers failed to make repayment of certain amount of money on the loan from MCI.'' Without this context, it is hard to determine that Ebbers should be the recipient of the loan.
    
    \item \textit{Lack of knowledge to obtain exact boundary of the argument span.} For example, in ``Negotiations between Washington and Pyongyang on their nuclear dispute have been set for April 23 in Beijing ...'', for the \textsc{Entity} role, two argument spans should be extracted (``Washington'' and ``Pyongyang''). While our framework predicts the entire ``Washington and Pyongyang'' as the argument span. Although there's an overlap between the prediction and gold-data, the model gets no credit for it.
    
    \item \textit{Data and lexical sparsity.} In the following two examples, our model fails to detect the triggers of type \textsc{End-Position}.
    ``Minister Tony Blair said \textbf{ousting} Saddam Hussein now was key to solving similar crises.''
    ``There's no indication if Erdogan would \textbf{purge} officials who opposed letting in the troops.''
    It's partially due to they not being seen during training as triggers.  ``ousting'' is a rare word and is not in the tokenizers' vocabulary. Purely inferring from the sentence context is hard to make the correct prediction.
\end{itemize}

\section{Related Work}

\paragraph{Event Extraction}
Most event extraction research has focused on the 2005 Automatic Content Extraction (ACE) sentence-level event task~\cite{walker2006ace}. In recent years, continuous representations from convolutional neural networks~\cite{nguyen2015event,chen-etal-2015-event} and recurrent neural networks~\cite{nguyen-etal-2016-joint} have been proved to help substantially for pipeline-based classifiers by automatically extracting features.
To mitigate the effect of error propagation, joint models have been proposed for event extraction.
\newcite{yang-mitchell-2016-joint} consider structural dependencies between events and entities, which requires heavy feature engineering to capture discriminative information.
\newcite{nguyen2019one} propose a multitask model that performs entity recognition, trigger detection and argument role prediction by sharing BiGRU hidden representations.
\newcite{zhangjunchi2019extracting} utilize a neural transition-based extraction framework~\cite{zhang2011syntactic}, which requires specially designed transition actions. It still requires recognizing entities during decoding, though entity recognition and argument role prediction are done jointly.

These methods generally perform {\bf trigger detection $\rightarrow$ entity recognition $\rightarrow$ argument role assignment} during decoding. Different from the works above, our framework completely bypasses the entity recognition stage (thus no annotation resources for NER needed), and directly tackles event argument extraction.
Also related to our work includes DYGIE++~\cite{wadden-etal-2019-entity} -- it models the entity/argument spans (with start and end offset) instead of labeling with the BIO scheme. Different from our work, its learned span representations are later used to predict the entity/argument type. While our QA model directly extracts the spans for certain argument role types.
Contextualized representations produced by pre-trained language models~\cite{peters-etal-2018-deep,devlin-etal-2019-bert} have been shown to be helpful for event extraction~\cite{zhangtongtao2019joint,wadden-etal-2019-entity} and question answering~\cite{rajpurkar-etal-2016-squad}. The attention mechanism helps capture relationships between tokens in the question and input sequence tokens. We use BERT in our framework for capturing these semantic relationships.

\paragraph{Machine Reading Comprehension (MRC)} 
Span-based MRC tasks involve extracting a span from a paragraph~\cite{rajpurkar-etal-2016-squad} or multiple paragraphs~\cite{joshi-etal-2017-triviaqa, kwiatkowski2019natural}.
Recently, there have been explorations on formulating NLP tasks as a question answering problem.
\newcite{mccann2018natural} proposes natural language decathlon challenge (decaNLP), which consists of ten tasks (e.g., machine translation, summarization, question answering). They cast all tasks as question answering over a context and propose a general model for this.
In the information extraction literature, \newcite{levy-etal-2017-zero} propose the zero-shot relation extraction task and reduce the task to answering crowd-sourced reading comprehension questions.
\newcite{li-etal-2019-entity} casts entity-relation extraction as a multi-turn question answering task. Their questions lack diversity and naturalness. For example for the \textsc{PART-WHOLE} relation, the template question is ``find Y that belongs to X'', where X is instantiated with the pre-given entity.
The follow-up work for named entity recognition from \newcite{li2019unified} propose better query strategies incorporating synonyms and examples.
Different from the works above, we focus on the more complex event extraction task, which involves both trigger detection and argument extraction. Our generated questions for extracting event arguments are somewhat more natural (incorporating descriptions from annotation guidelines) and leverage trigger information.

\paragraph{Question Generation} 
To generate question templates 2\&3 (Type + Role question and annotation guideline based question) which are more natural, we draw insights from the literature of automatic rule-based question generation~\cite{heilman2010good}.
\newcite{heilman2011automatic} propose to use linguistically motivated rules for WH word (question phrase) selection. In their more general case of question generation from sentences, answer phrases can be noun phrases, prepositional phrases, or subordinate clauses. Complicated rules are designed with help from the superTagger~\cite{ciaramita2006broad}.
In our case, event arguments are mostly noun phrases and the rules are simpler -- ``who'' for person, ``where'' for place and ``what'' for all other types of entities. We sample around 10 examples from the development set to determine the entity type of each argument role.
In the future, it will be interesting to investigate how to utilize machine learning-based question generation methods~\cite{du-etal-2017-learning}. They would be more beneficial for the setting where the schema/ontology contains a large number of argument types.

\section{Conclusion}
In this paper, we introduce a new paradigm for event extraction based on question answering.
We investigate how the question generation strategies affect the performance of our framework on both trigger detection and argument span extraction, and find that more natural questions lead to better performance.
Our framework outperforms prior works on the ACE 2005 benchmark, and is capable of extracting event arguments of roles not seen at training time.
For future work, it would be interesting to try incorporating broader context (e.g., paragraph/document-level context~\cite{ji-grishman-2008-refining, huang-riloff-2011-peeling, du-cardie-2020-document} in our methods to improve the accuracy of the predictions.

\section*{Acknowledgments}

We thank the anonymous reviewers and Heng Ji for helpful suggestions. This research is based on work supported in part by DARPA LwLL Grant FA8750-19-2-0039.

\bibliography{EMNLP2020}
\bibliographystyle{acl_natbib}

\onecolumn
\appendix
\section{Questions Based on Annotation Guidelines}
\label{sec:appendix}
Questions based on annotation guidelines for each argument role.

\begin{table}[!h]
\small
\centering
\begin{tabular}{l|l|l}
\toprule
\textbf{Event Type} & \textbf{Argument Role} & \textbf{Question}  \\ \midrule
\multirow{2}{*}{Business.Declare-Bankruptcy}    & Org           & What declare bankruptcy?                                  \\
                                                & Place         & Where the event takes place?                              \\ \midrule
\multirow{2}{*}{Business.End-Org}               & Org           & What is ended?                                            \\
                                                & Place         & Where the event takes place?                              \\ \midrule
Business.Merge-Org                              & Org           & What is merged?                                           \\ \midrule
\multirow{3}{*}{Business.Start-Org}             & Org           & What is started?                                          \\
                                                & Place         & Where the event takes place?                              \\
                                                & Agent         & Who is the founder?                                       \\ \midrule
\multirow{5}{*}{Conflict.Attack}                & Place         & Where the event takes place?                              \\
                                                & Target        & Who is the target?                                        \\
                                                & Attacker      & Who is the attacking agent?                               \\
                                                & Instrument    & What is the instrument used?                              \\
                                                & Victim        & Who is the victim?                                        \\ \midrule
\multirow{2}{*}{Conflict.Demonstrate}           & Entity        & Who is demonstrating agent?                               \\
                                                & Place         & Where the event takes place?                              \\ \midrule
\multirow{2}{*}{Contact.Meet}                   & Entity        & Who is meeting?                                           \\
                                                & Place         & Where the event takes place?                              \\ \midrule
\multirow{2}{*}{Contact.Phone-Write}            & Entity        & Who is communicating agents?                              \\
                                                & Place         & Where the event takes place?                              \\ \midrule
\multirow{2}{*}{Justice.Acquit}                 & Defendant     & Who is the defendant?                                     \\
                                                & Adjudicator   & What is the judge?                                        \\ \midrule
\multirow{3}{*}{Justice.Appeal}                 & Adjudicator   & What is the judge?                                        \\
                                                & Plaintiff     & What is the plaintiff?                                    \\
                                                & Place         & Where the event takes place?                              \\ \midrule
\multirow{3}{*}{Justice.Arrest-Jail}            & Person        & Who is jailed?                                            \\
                                                & Agent         & Who is the jailor?                                        \\
                                                & Place         & Where the event takes place?                              \\ \midrule
\multirow{4}{*}{Justice.Charge-Indict}          & Adjudicator   & What is the judge?                                        \\
                                                & Defendant     & Who is the defendant?                                     \\
                                                & Prosecutor    & Who is the prosecuting agent?                             \\
                                                & Place         & Where the event takes place?                              \\ \midrule
\multirow{3}{*}{Justice.Convict}                & Defendant     & Who is the defendant?                                     \\
                                                & Adjudicator   & What is the judge?                                        \\
                                                & Place         & Where the event takes place?                              \\ \midrule
\multirow{3}{*}{Justice.Execute}                & Place         & Where the event takes place?                              \\
                                                & Agent         & Who carry out the execution?                              \\
                                                & Person        & Who was executed?                                         \\ \midrule
\multirow{3}{*}{Justice.Extradite}              & Origin        & What is original location of the person being extradited? \\
                                                & Destination   & Where the person is extradited to?                        \\
                                                & Agent         & Who is the extraditing agent?                             \\ \midrule
\multirow{3}{*}{Justice.Fine}                   & Entity        & What is fined?                                            \\
                                                & Adjudicator   & What is the judge?                                        \\
                                                & Place         & Where the event takes place?                              \\ \midrule
\multirow{3}{*}{Justice.Pardon}                 & Adjudicator   & What is the judge?                                        \\
                                                & Place         & Where the event takes place?                              \\
                                                & Defendant     & Who is the defendant?                                     \\ \midrule
\multirow{3}{*}{Justice.Release-Parole}         & Entity        & Who will do the release?                                  \\
                                                & Person        & Who is released?                                          \\
                                                & Place         & Where the event takes place?                              \\ \midrule
\multirow{3}{*}{Justice.Sentence}               & Defendant     & Who is the defendant?                                     \\
                                                & Adjudicator   & What is the judge?                                        \\
                                                & Place         & Where the event takes place?                              \\   \bottomrule
\end{tabular}
\end{table}

\newpage
\begin{table}[h]
\small
\centering
\begin{tabular}{l|l|l}
\toprule
\multirow{4}{*}{Justice.Sue}                    & Plaintiff     & What is the plaintiff?                                    \\
                                                & Defendant     & Who is the defendant?                                     \\
                                                & Adjudicator   & What is the judge?                                        \\
                                                & Place         & Where the event takes place?                              \\ \midrule
\multirow{4}{*}{Justice.Trial-Hearing}          & Defendant     & Who is the defendant?                                     \\
                                                & Place         & Where the event takes place?                              \\
                                                & Adjudicator   & What is the judge?                                        \\
                                                & Prosecutor    & Who is the prosecuting agent?  \\ \midrule
\multirow{2}{*}{Life.Be-Born}   & Place         & Where the event takes place?  \\ 
                                                & Person        & Who is born?                                              \\ \midrule
\multirow{4}{*}{Life.Die}                       & Victim        & Who died?                                                 \\
                                                & Agent         & Who is the killer?                                        \\
                                                & Place         & Where the event takes place?                              \\
                                                & Instrument    & What is the instrument used?                              \\ \midrule
\multirow{2}{*}{Life.Divorce}                   & Person        & Who are divorced?                                         \\
                                                & Place         & Where the event takes place?                              \\ \midrule
\multirow{4}{*}{Life.Injure}                    & Victim        & Who is victim?                                            \\
                                                & Agent         & Who is the attacking agent?                               \\
                                                & Place         & Where the event takes place?                              \\
                                                & Instrument    & What is the instrument used?                              \\ \midrule
\multirow{2}{*}{Life.Marry}                     & Person        & Who are married?                                          \\
                                                & Place         & Where the event takes place?                              \\ \midrule
\multirow{5}{*}{Movement.Transport}             & Vehicle       & What is the vehicle used?                                 \\
                                                & Artifact      & What is being transported?                                \\
                                                & Destination   & Where the transporting is directed?                       \\
                                                & Agent         & Who is responsible for the transport event?               \\
                                                & Origin        & Where the transporting originated?                        \\ \midrule
\multirow{3}{*}{Personnel.Elect}                & Person        & Who is elected?                                           \\
                                                & Entity        & Who voted?                                                \\
                                                & Place         & Where the event takes place?                              \\ \midrule
\multirow{3}{*}{Personnel.End-Position}         & Entity        & Who is the employer?                                      \\
                                                & Person        & Who is the employee?                                      \\
                                                & Place         & Where the event takes place?                              \\ \midrule
\multirow{2}{*}{Personnel.Nominate}             & Person        & Who is nominated?                                         \\
                                                & Agent         & Who is the nominating agent?                              \\ \midrule
\multirow{3}{*}{Personnel.Start-Position}       & Person        & Who is the employee?                                      \\
                                                & Entity        & Who is the employer?                                      \\
                                                & Place         & Where the event takes place?                              \\ \midrule
\multirow{4}{*}{Transaction.Transfer-Money}     & Giver         & Who is the donating agent?                                \\
                                                & Recipient     & Who is the recipient?                                     \\
                                                & Beneficiary   & Who benefits from the transfer?                           \\
                                                & Place         & Where the event takes place?                              \\ \midrule
\multirow{5}{*}{Transaction.Transfer-Ownership} & Buyer         & Who is the buying agent?                                  \\
                                                & Artifact      & What was bought?                                          \\
                                                & Seller        & Who is the selling agent?                                 \\
                                                & Place         & Where the event takes place?                              \\
                                                & Beneficiary   & Who benefits from the transaction?   \\  \bottomrule
\end{tabular}
\end{table}




\end{document}